\def\year{2021}\relax
\documentclass[letterpaper]{article} 
\usepackage{aaai21}  
\usepackage{times}  
\usepackage{helvet} 
\usepackage{courier}  
\usepackage[hyphens]{url}  
\usepackage{graphicx} 
\usepackage{natbib}  
\usepackage{caption} 
\usepackage{booktabs}
\usepackage{xspace}
\usepackage{xcolor}
\usepackage{tabu}

\usepackage{tikz}
\usetikzlibrary{tikzmark}

\newcommand{\dataset}{RISE\xspace}
\newcommand{\datasize}{12,567\xspace}

\newcommand{\quo}[1]{``{#1}''}
\newcommand{\etal}{et al.\xspace}

\title{Project \dataset: Recognizing Industrial Smoke Emissions}
\author{
	Yen-Chia Hsu\textsuperscript{\rm 1}, Ting-Hao (Kenneth) Huang\textsuperscript{\rm 2}, Ting-Yao Hu\textsuperscript{\rm 1}, Paul Dille\textsuperscript{\rm 1}, Sean Prendi\textsuperscript{\rm 1}, \\ Ryan Hoffman\textsuperscript{\rm 1}, Anastasia Tsuhlares\textsuperscript{\rm 1}, Jessica Pachuta\textsuperscript{\rm 1}, Randy Sargent\textsuperscript{\rm 1}, Illah Nourbakhsh\textsuperscript{\rm 1}
	\\
}
\affiliations{
	
	\textsuperscript{\rm 1}Carnegie Mellon University, Pittsburgh, PA, USA \\ \textsuperscript{\rm 2}Pennsylvania State University, State College, PA, USA\\
	
	\{yenchiah, tingyaoh, pdille, sprendi, ryanhoff, anastt, jpachuta, rsargent, illah\}@andrew.cmu.edu, txh710@psu.edu
	
}

\begin{document}
	
	\frenchspacing
	\maketitle
	
	\begin{abstract}
		Industrial smoke emissions pose a significant concern to human health.
		Prior works have shown that using Computer Vision (CV) techniques to identify smoke as visual evidence can influence the attitude of regulators and empower citizens to pursue environmental justice.
		However, existing datasets are not of sufficient quality nor quantity to train the robust CV models needed to support air quality advocacy.
		We introduce \textbf{\underline{\dataset}}, the first large-scale video dataset for \textbf{\underline{R}ecognizing \underline{I}ndustrial \underline{S}moke \underline{E}missions}.
		We adopted a citizen science approach to collaborate with local community members to annotate whether a video clip has smoke emissions.
		Our dataset contains \datasize clips from 19 distinct views from cameras that monitored three industrial facilities.
		These daytime clips span 30 days over two years, including all four seasons.
		We ran experiments using deep neural networks to establish a strong performance baseline and reveal smoke recognition challenges.
		Our survey study discussed community feedback, and our data analysis displayed opportunities for integrating citizen scientists and crowd workers into the application of Artificial Intelligence for Social Impact.
	\end{abstract}
	
	\begin{table*}[t]
		\centering
		\footnotesize
		\begin{tabular}{lccccccccc}
			\toprule
			& \begin{tabular}{@{}c@{}} \# of \\ views \end{tabular}
			& \begin{tabular}{@{}c@{}@{}} \# of \\ labeled \\ clips \end{tabular}
			& \begin{tabular}{@{}c@{}@{}} \# of \\ frames \\ (images) \end{tabular}
			& \begin{tabular}{@{}c@{}@{}} Average \\ \# of frames \\ per clip \end{tabular}
			& \begin{tabular}{@{}c@{}@{}} Ratio of \\ smoke \\ frames \end{tabular}
			&  \begin{tabular}{@{}c@{}} Has \\ temporal \\ data? \end{tabular}
			& \begin{tabular}{@{}c@{}@{}} Has \\ context? \end{tabular}
			& \begin{tabular}{@{}c@{}@{}} Is from \\ industrial \\ sources? \end{tabular}
			& \begin{tabular}{@{}c@{}@{}} Appearance \\ change level \end{tabular} \\
			\midrule
			This Work & 19 & \datasize & 452,412 & 36 & 41\% & yes & yes & yes & high \\
			\midrule
			Bugaric \etal~\citeyear{bugaric-2014-adaptive} & 10 & 10 & 213,909 & 21,391 & 100\% & yes & yes & no & low \\
			Ko \etal~\citeyear{ko-2013-spatiotemporal} & 16 & 16 & 43,090 & 1,514 & 37\% & yes & yes & no & low \\
			Dimitropoulos \etal~\citeyear{dimitropoulos-2014-spatio} & 22 & 22 & 17,722 & 806 & 56\% & yes & yes & no & low \\
			Toreyin \etal~\citeyear{toreyin-2005-wavelet} & 21 & 21 & 18,031 & 820 & 98\% & yes & yes & no & low \\
			Filonenko \etal~\citeyear{filonenko-2017-smoke}* & ... & 396 & 100,968 & 255 & 61\% & yes & no & no & low \\
			\midrule
			Xu \etal~\citeyear{xu-2019-video} & ... & ... & 5,700 & ... & 49\% & no & yes & no & low \\
			Xu \etal~\citeyear{xu-2019-adversarial} & ... & ... & 3,578 & ... & 100\% & no & yes & no & low \\
			Xu \etal~\citeyear{xu-2017-deep} & ... & ... & 10,000 & ... & 50\% & no & yes & no & low \\
			Ba \etal~\citeyear{ba-2019-smokenet}*$\dagger$ & ... & ... & 6,225 & ... & 16\% & no & no & unknown & medium \\
			Lin \etal~\citeyear{lin-2017-smoke}* & ... & ... & 16,647 & ... & 29\% & no & no & no & low \\
			Yuan \etal~\citeyear{yuan-2019-convolutional}* & ... & ... & 24,217 & ... & 24\% & no & no & no & low \\
			\bottomrule
		\end{tabular}
		\caption{Comparison of publicly available datasets for recognizing smoke. This table does not include unlabeled or synthetic data. Symbol \quo{...} means not applicable, where the corresponding datasets are image-based. The * symbol means that the dataset provides labels on image patches that do not provide sufficient context for the surroundings. The $\dagger$ symbol means that the dataset uses satellite images, and therefore the sources of smoke emissions are unidentifiable. We used the \texttt{ffprobe} command in FFmpeg to count video frames. Some datasets treat steam as smoke, and we counted the number of smoke frames within these datasets based only on videos that do not involve steam. Moreover, some datasets contain videos for fire detection, and we did not count them in this table, since fire detection was not our focus in this research.}
		\label{tab:dataset-comparison}
	\end{table*}
	
	\section{Introduction}
	
	Air pollution has been associated with adverse impacts on human health~\cite{kampa-2008,pope-2006,dockery-1993}.
	According to the United States Environmental Protection Agency (US EPA), air pollutants emitted from industrial sources pose a significant concern\footnote{Link to US EPA Air Pollution Sources:\\ \url{https://www.epa.gov/stationary-sources-air-pollution}}.
	Currently, citizens who wish to advocate for better air quality rely on a manual approach (US EPA Visual Opacity Reading\footnote{Link to US EPA Visual Opacity Reading:\\ \url{https://www.epa.gov/emc/method-9-visual-opacity}}) to determine if smoke emissions violate the permit issued to the facility.
	This laborious approach requires certification every six months and involves taking multiple field measurements.
	Prior works have shown that using Computer Vision to identify industrial smoke emissions automatically can empower citizens to pursue environmental justice and urge regulators to respond to local concerns publicly~\cite{hsu-2016,hsu-2017}.
	This type of data-driven evidence, when integrated with community narratives, is essential for citizens to make sense of environmental issues and take action~\cite{ottinger-2017-making,hsu-2017}.
	
	\begin{figure}[t]
		\centering
		\includegraphics[width=0.7\columnwidth]{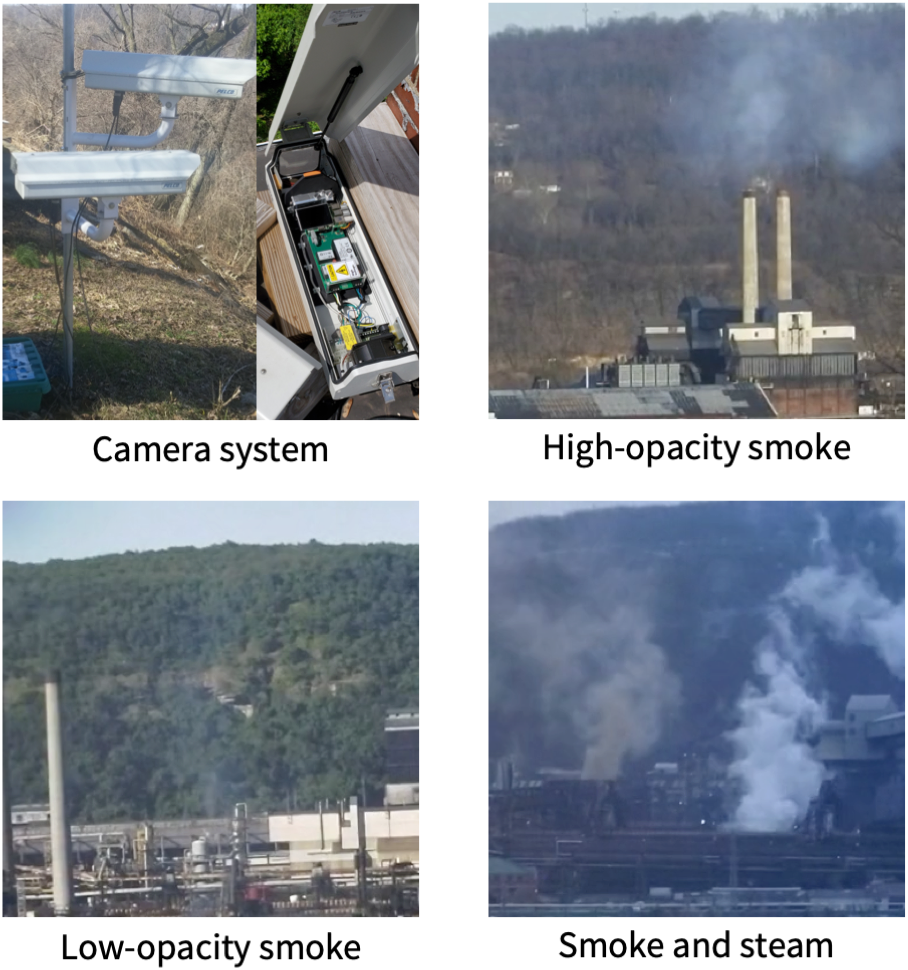}
		\caption{Dataset samples and the deployed camera system.}
		\label{fig:camera}
	\end{figure}
	
	However, it is tough to collect large-scale data required to build a practical industrial smoke recognition model.
	Recent state-of-the-art models based on deep learning typically have millions of parameters and are data-hungry.
	Training these models with insufficient data can lead to overfitting.
	Table~\ref{tab:dataset-comparison} lists publicly available datasets for smoke recognition.
	Compared to object and action recognition datasets, such as ImageNet (14M images, \citeauthor{russakovsky-2015-imagenet} \citeyear{russakovsky-2015-imagenet}) and Kinetics (650K videos, \citeauthor{kay-2017-kinetics} \citeyear{kay-2017-kinetics}), existing ones for smoke recognition are small.
	There have been successes in prior works using deep learning for smoke recognition~\cite{yuan-2019-convolutional,ba-2019-smokenet,xu-2017-deep,xu-2019-adversarial,xu-2019-video,filonenko-2017-smoke,liu-2019-dual,yin-2017-deep,yang-2018-combining,lin-2019-smoke,hu-2018-real,zhang-2018-wildland}, but these models were trained and evaluated on relatively small datasets.
	In response to data sparsity, some works have attempted to create artificial data, where smoke with a transparent background were synthesized with various views~\cite{yuan-2019-deep,zhang-2018-wildland,xu-2019-adversarial}.
	But such synthetic data cannot capture the rich behavior and appearance of smoke under real-world conditions, such as changing weather.
	Moreover, existing smoke-recognition datasets lack sufficient quality for our task.
	No existing ones contain smoke emissions from industrial sources.
	Most of them are from burning materials ({\em e.g.}, in labs) and fire events ({\em e.g.}, wildfire), which have low temporal appearance changes.
	Thirty-six percent of them are imbalanced (ratio of smoke frames higher than 80\% or less than 20\%).
	The average number of frames per labeled clip is high in them, indicating weak temporal localization.
	
	This paper introduces \textbf{\underline{\dataset}}, the first large-scale video dataset for \textbf{\underline{R}ecognizing \underline{I}ndustrial \underline{S}moke \underline{E}missions}.
	We built and deployed camera systems to monitor activities of various industrial coke plants in Pittsburgh, Pennsylvania (Figure~\ref{fig:camera}).
	We collaborated with air quality grassroots communities to install the cameras, which capture an image about every 10 seconds.
	These images were streamed back to our servers, stitched into panoramas, and stacked into timelapse videos.
	From these panorama videos, we cropped clips based on domain knowledge about where smoke emissions frequently occur.
	Finally, these clips were labeled as yes/no (``whether smoke emissions exist'') by volunteers using a web-based tool that we developed (Figure~\ref{fig:ui-smoke} and~\ref{fig:ui-smoke-expert}).
	
	Our contributions include: (1) building a dataset to facilitate using AI for social impact, (2) showing a way to empower citizens through AI research, and (3) making practical design challenges explicit to others.
	\dataset contains clips from 30 days spanning four seasons over two years, all taken in daylight.
	The labeled clips have 19 views cropped from three panoramas taken by cameras at three locations.
	The dataset covers various characteristics of smoke, including opacity and color, under diverse weather and lighting conditions.
	Moreover, the dataset includes distractions of various types of steam, which can be similar to smoke and challenging to distinguish.
	We use the dataset to train an I3D ConvNet~\cite{carreira-2017-quo} as a strong baseline benchmark.
	We compare the labeling quality between citizen scientists and online crowd workers on Amazon Mechanical Turk (MTurk).
	Using a survey study, we received community feedback.
	The code and dataset\footnote{Link to the Dataset and Code:\\ \url{https://github.com/CMU-CREATE-Lab/deep-smoke-machine}} are open-source, as is the video labeling system\footnote{Link to the Video Labeling System:\\ \url{https://github.com/CMU-CREATE-Lab/video-labeling-tool}}.
	
	\section{Related Work}
	
	A set of prior work relied on physics or hand-crafted features to recognize smoke.
	Kopilovic \etal~(\citeyear{kopilovic-2000-application}) computed the entropy of optical flow to identify smoke.
	Celik \etal~(\citeyear{celik-2007-fire}) used color to define smoke pixels.
	Toreyin \etal~(\citeyear{toreyin-2005-wavelet}) combined background subtraction, edge flickering, and texture analysis.
	Lee \etal~(\citeyear{lee-2012-smoke}) used change detection to extract candidate regions, computed features based on color and texture, and trained a support vector machine classifier using these features.
	Tian \etal~(\citeyear{tian-2015-single}) presented a physical-based model and used sparse coding to extract reliable features for single-image smoke detection.
	Gubbi \etal~(\citeyear{gubbi-2009-smoke}) and Calderara \etal~(\citeyear{calderara-2008-smoke}) applied texture descriptors (such as a wavelet transform) on small image blocks to obtain feature vectors and train a classifier using these features.
	These works relied on heuristics to tune the hand-crafted features, which makes it tough to be robust~\cite{hsu-2016}.
	
	Another set of prior work developed or enhanced various deep learning architectures for smoke recognition.
	For example, Yuan \etal~(\citeyear{yuan-2019-deep}) trained two encoder-decoder networks to focus on global contexts and fine details, respectively, for smoke region segmentation.
	Hu and Lu~(\citeyear{hu-2018-real}) trained spatial-temporal ConvNets using multi-task learning.
	Liu \etal~(\citeyear{liu-2019-dual}) classified smoke by fusing ResNet and ConvNet trained with the original RGB and Dark Channel images~\cite{he-2010-single}, respectively.
	Other work applied or enhanced object detectors, such as SSD~\cite{liu-2016-ssd}, MS-CNN~\cite{cai-2016-unified}, Faster R-CNN~\cite{ren-2015-faster}, and YOLO~\cite{redmon-2017-yolo9000}, to identify regions that have smoke~\cite{xu-2019-adversarial,zhang-2018-wildland,yang-2018-combining,lin-2019-smoke}.
	These works were evaluated on small datasets (Table~\ref{tab:dataset-comparison}), and none of them collaborated with local communities in air quality advocacy.

	We employed citizen science~\cite{shirk-2012-public} to simultaneously empower laypeople and build a large dataset.
	Citizen science has been successfully applied in science projects, especially when the research scale makes it infeasible for experts to tackle alone.
	For example, PlantNet invites volunteers to submit images to develop a plant species classification model~\cite{joly2016crowdsourcing}.
	In PlantNet, experts define research questions and invite citizens to participate.
	On the other hand, we utilized data-driven evidence to address concerns of local communities.
	Specifically, we applied a civic-oriented framework, \textit{Community Citizen Science}~\cite{hsu2020human}, to empower citizens affected by industrial pollution to advocate for better air quality.
	
	\section{\dataset Dataset}
	
	\begin{table}[t]
		\centering
		\footnotesize
		\begin{tabular}{lcc}
			\toprule
			& \# of labeled clips & Ratio \\
			\midrule
			Has smoke & 5,090 & 41\% \\
			\midrule
			Winter (Dec to Feb) & 7,292 & 58\% \\
			Spring (Mar to May) & 1,057 & 8\% \\
			Summer (Jun to Aug) & 2,999 & 24\% \\
			Fall (Sep to Nov) & 1,219 & 10\% \\
			\midrule
			6 am to 10 am & 4,001 & 32\% \\
			11 am to 3 pm & 6,071 & 48\% \\
			4 pm to 8 am & 2,495 & 20\% \\
			\bottomrule
		\end{tabular}
		\caption{The number and ratio of video clips for all 19 camera views filtered by various temporal conditions.}
		\label{tab:dataset}
	\end{table}
	
	\begin{figure*}[t]
		\centering
		\includegraphics[width=0.95\textwidth]{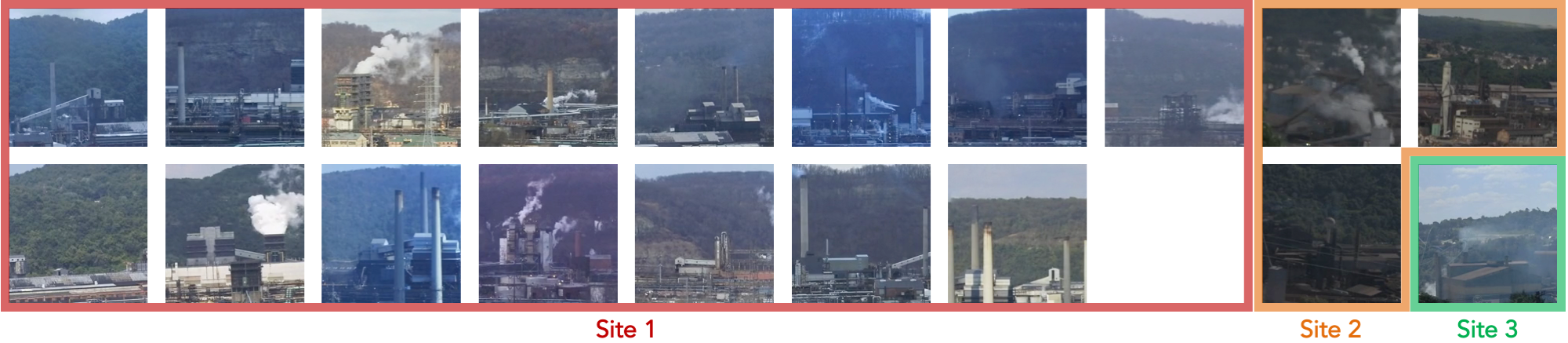}
		\caption{All views of videos in the \dataset dataset. The rightmost four views are from different sites pointing at another facility.}
		\label{fig:19-views}
	\end{figure*}

	Unlike static data annotation projects, \dataset is powered by local communities, where multiple factors can influence the data distribution (explained in the Discussion section).
	The \dataset dataset has \datasize labeled clips from industrial sources, including those emitted from stacks and escaped from facilities.
	Each clip has 36 frames (resolution 180x180 pixels), representing about six minutes in real-world time.
	These clips contain 19 views (Figure~\ref{fig:19-views}), where 15 are cropped from the panorama timelapse at one site, and four from two other sites.
	These clips span 30 days from two years, including all four seasons.
	They were taken in the daytime and include different weather and lighting conditions.
	Using weather archives and checking the videos, we manually selected these 30 days to include various conditions, balance the number of videos that have smoke, skip days when cameras were down, and add hard cases ({\em e.g.}, snow days).
	Of the 30 days, 20 days are cloudy, four days are fair, six days are a mixture of fair and cloudy, six days have light rain, four days have light snow, two days are windy, and one day has a thunderstorm.
	Using domain knowledge, we manually picked the 19 views to cover the locations of smoke emission sources.
	Since the length of day is different across seasons, fewer videos are in the morning (6 am to 10 am) and evening (4 pm to 8 pm).
	Table~\ref{tab:dataset} summarizes the dataset.
	
	\subsection{System for Data Collection}
	
	We built and deployed camera systems to monitor pollution sources (Figure~\ref{fig:camera}).
	Each system had a Nikon 1 J5 camera controlled by Raspberry Pi.
	It had a heater and cooler to control temperature and a servo motor to enable remote power cycling.
	The cost per camera was about \$2,000 US dollars.
	The camera took a photo about every 5 to 10 seconds, and the photos were streamed to a server for stitching and stacking into panorama timelapse videos.
	Areas in videos that look inside house windows were cropped or blocked.
	
	Deploying cameras relied heavily on citizen engagement.
	To build relationships and mutual trust between our organization and the community, we canvassed the local region, met with affected residents, and attended local events ({\em e.g.}, community meetings) to explain our mission.
	We installed the equipment and offered complimentary internet service to those residents willing to host cameras.
	
	We chose to use timelapse instead of video capturing due to real-world deployment constraints.
	We deployed 10 camera systems.
	To zoom into locations of emission sources, our camera captured each photo in 5K resolution.
	Using video capturing meant streaming 10 high-resolution videos back to our server, which would be impractical since many camera locations did not have the high-speed internet infrastructure to enable such large data-transmission bandwidth. Moreover, our 10 cameras produced about 600 GB data per day.
	Recording videos in 20 frames per second would generate 120 TB data per day, which was beyond our system's storage space and computational power.
	
	\begin{figure}[t]
		\centering
		\includegraphics[width=0.88\columnwidth]{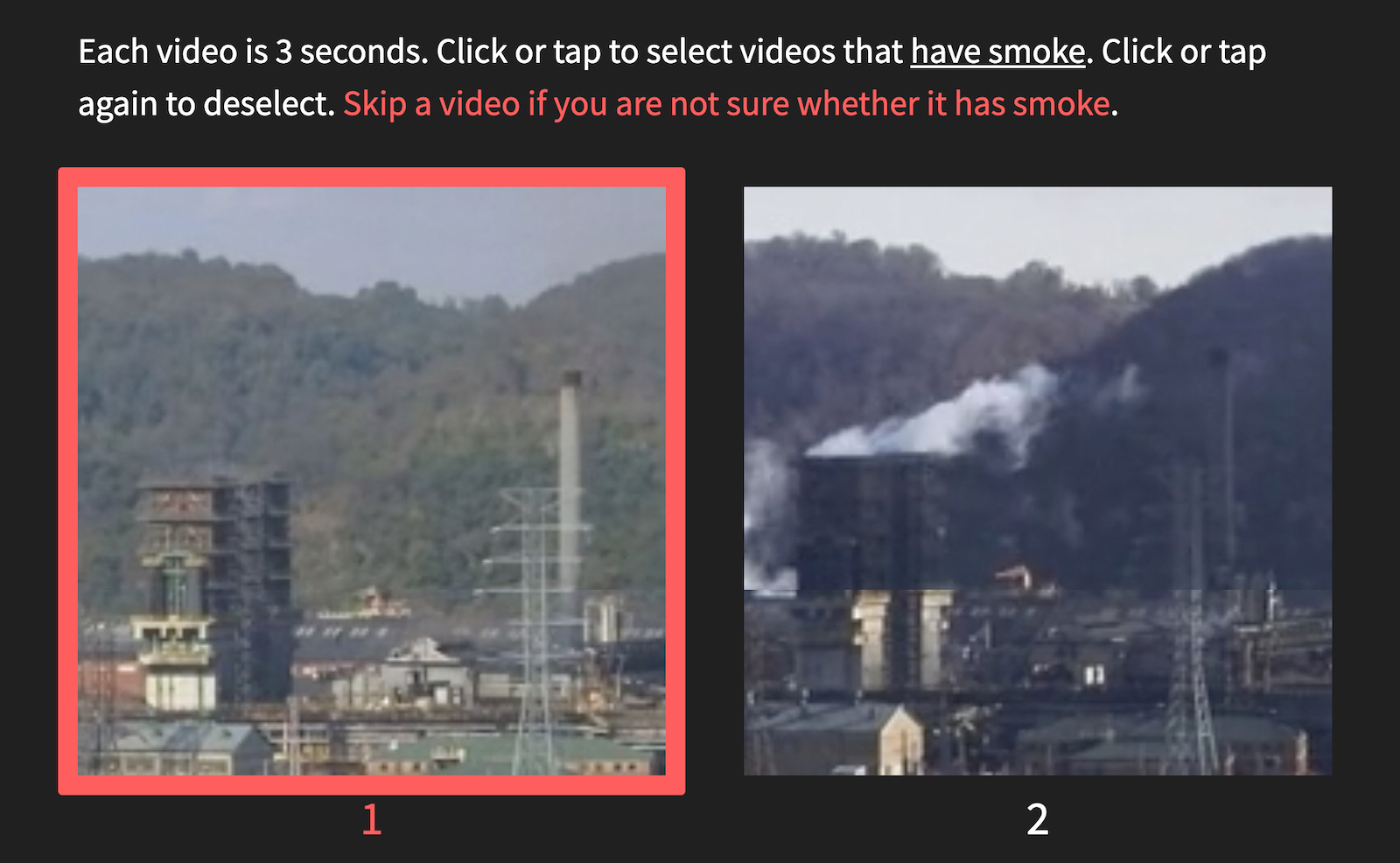}
		\caption{The individual mode of the smoke labeling system. Users can scroll the page and click or tap on the video clips to indicate that the video has smoke with a red border-box.}
		\label{fig:ui-smoke}
		\vspace{4mm}
		\centering
		\includegraphics[width=0.88\columnwidth]{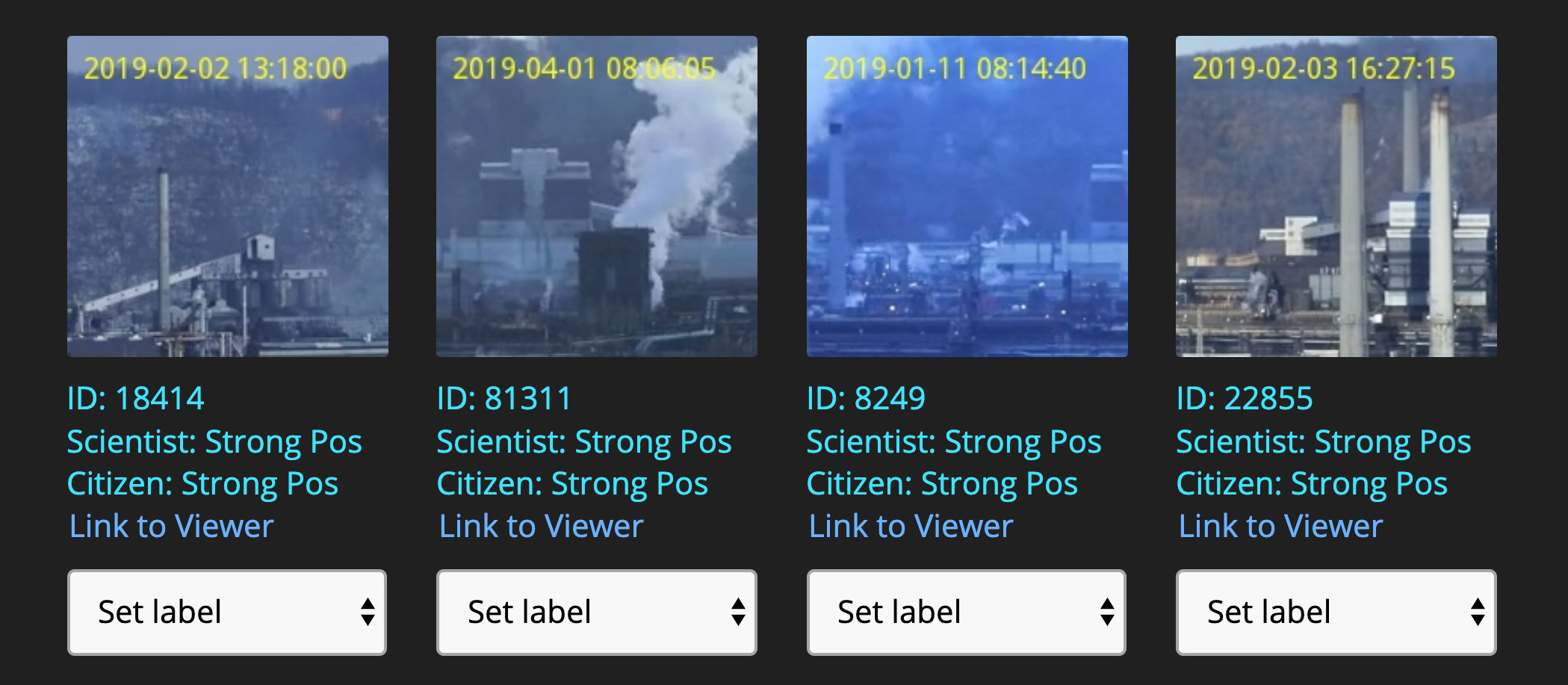}
		\caption{The collaborative mode of the smoke labeling system. Researchers can confirm citizen-provided labels by using the dropdown menu below each video.}
		\label{fig:ui-smoke-expert}
	\end{figure}
	
	\subsection{System for Data Annotation}
	
	We applied citizen science~\cite{shirk-2012-public,irwin-2002-citizen} to engage volunteers in this research and develop a web-based tool that allows volunteers to annotate data.
	Citizen science opens opportunities to collaborate with residents and advocacy groups who have diverse expertise in local concerns.
	Through two workshops with air quality advocates, three presentations during community events, and two guest lectures at universities, we recruited volunteers to help label smoke emissions.
	The design and use of this tool were iteratively refined based on community feedback.
	For example, many volunteers mislabeled smoke initially during the workshops, suggesting that the task can be challenging for those unfamiliar with the US EPA Visual Opacity Reading.
	Thus, we implemented an interactive tutorial to introduce the task with step-by-step guidance.
	Users are first presented with simplified tasks that explain concepts, such as the characteristics of smoke and the difference between smoke and steam.
	After performing each small task, the system shows the answer, explanation, and the location of smoke.
	
	In our tool, there are two modes for data labeling.
	The first one (individual mode) asks volunteers or researchers to label 16 randomly-chosen video clips at once (Figure~\ref{fig:ui-smoke}).
	Users can scroll the interface and click or tap on clips to indicate the presence of smoke emissions.
	The second one (collaborative mode) asks researchers to confirm the labels contributed by volunteers, where citizens' answers are shown as prior information, and the researcher can choose to agree with or override these answers (Figure~\ref{fig:ui-smoke-expert}).
	This mode is designed to reduce researcher's mental load and increase the speed of gathering labels at the early stage of system deployment when the number of users is few.
	
	Instead of labeling opacity, we decided to perform the simpler task of annotating whether smoke exists.
	Negative labels mean no smoke, and positive labels mean smoke appears at some time.
	This weak labeling approach is more straightforward for laypeople, which enabled broader citizen participation.
	When labeling, we referred to the 16 clips provided by the system as a page.
	For quality control, we randomly inserted four gold standards (labeled by a researcher).
	The system accepts a page if the volunteer answers all the gold standards correctly.
	Otherwise, the page is discarded.
	At least one negative and one positive sample are included in the gold standards to prevent uniform guessing.
	Also, each clip was reviewed by at least two volunteers (identified using Google Analytics tracker).
	If their answers did not agree, a third volunteer was asked to review the clip.
	The system takes the majority vote of the answers as the final label.
	
	\subsection{Analysis of User Contribution and Data Quality}
	
	\begin{table*}[t]
		\centering
		\footnotesize
		\begin{tabular}{lccccc}
			\toprule
			User group
			& \# of users
			& \begin{tabular}{@{}c@{}} Page acceptance rate \\ $\forall$ group \end{tabular}
			& \begin{tabular}{@{}c@{}} Page acceptance rate \\ $\forall$ user \end{tabular}
			& \begin{tabular}{@{}c@{}} \# of accepted pages \\ $\forall$ group  \end{tabular}
			& \begin{tabular}{@{}c@{}} \# of accepted pages \\ $\forall$ user \end{tabular} \\
			\midrule
			Top Enthusiasts & 7 (12\%) & .86 & .76$\pm$.10 & 1,491 (86\%) & 213$\pm$328 \\
			Other Enthusiasts & ... & ... & ... & ... & ... \\
			Top Volunteers & 41 (68\%) & .69 & .74$\pm$.19 & 218 (13\%) & 5$\pm$5 \\
			Other Volunteers &  12 (20\%) & .26 & .28$\pm$.08 & 18 (1\%) &  2$\pm$1 \\
			All & 60 (100\%) & .81 & .65$\pm$.25 & 1,727 (100\%) & 29$\pm$125 \\
			\bottomrule
		\end{tabular}
		\caption{Analysis of volunteers (excluding the researchers) who contributed at least one page (with 16 clips) that passed the quality check and was accepted by the system. The format for the 4th and 6th columns is \quo{mean$\pm$standard deviation}.}
		\label{tab:user_group}
	\end{table*}
	
	The tool was launched in early February 2019.
	Through February 24, 2020 (12 months), we had \datasize labeled clips.
	Among them, 42\% (5,230) and 20\% (2,560) were labeled by researchers and citizens in the individual mode, and 38\% (4,777) were labeled in the collaborative mode.
	During the 12-month launch period, we attracted 60 volunteers who contributed at least one page (with 16 clips) that passed the system's quality check.
	Most of our volunteers received in-person training from a researcher in a 3-hour workshop.
	Of the entire labeled videos, only 193 (less than 2\%) had volunteer label disagreement.
	Table~\ref{tab:user_group} showed that 12\% of the volunteers contributed 86\% of the data (the top enthusiast group).
	Volunteers in this group had a higher acceptance rate ($\geq$ 0.5) and a higher number of accepted pages ($\geq$ the average of all participants).
	This skewed pattern of contribution is typical among citizen science projects such as Zooniverse (\url{https://zooniverse.org}), where many volunteers participate only a few times~\cite{sauermann2015crowd}.
	
	We compared the labels produced by citizens, researchers, and MTurk workers.
	For MTurk workers, we randomly sampled 720 clips from 10,625 ones that had been labeled by both citizens and researchers between February 2019 and November 2019.
	We then divided those clips into 60 tasks.
	For quality control, we added four randomly-sampled gold standards to each task.
	The user interface was identical to the one used for citizens.
	Differently, the interactive tutorial is required for MTurk workers before labeling.
	We first posted the tutorial task with 50 assignments to Amazon Mechanical Turk (\$1.50 per task).
	We then posted 60 labeling tasks, where each task collected five assignments from different workers.
	Only workers who finished the tutorial task could perform these labeling tasks.
	The estimated time to complete a labeling task was 90 seconds.
	We paid \$0.25 per labeling task, yielding an hourly wage of \$15.
	Fourteen workers were recruited and accomplished all tasks in about 12 hours.
	
	\begin{table}[t]
		\centering
		\footnotesize
		\begin{tabular}{lccc}
			\toprule
			User group & Precision & Recall & F-score\\
			\midrule
			Citizen & .98 & .83 & .90 \\
			Filtered MTurk workers & .94$\pm$.01 & .89$\pm$.01 & .91$\pm$.01 \\
			All MTurk workers & .93$\pm$.01 & .83$\pm$.01 & .88$\pm$.01 \\
			\bottomrule
		\end{tabular}
		\caption{The data quality (simulated 100 times) of citizens and MTurk workers, using researcher labels as the ground truth based on 720 labeled videos with 392 positive labels. Filtered MTurk workers' page acceptance rate is larger than 0.3. The reported format is \quo{mean$\pm$standard deviation.}}
		\label{tab:mturk_citizen_compare}
	\end{table}
	
	\begin{table}[t]
		\centering
		\footnotesize
		\begin{tabular}{lc}
			\toprule
			& Cohen's kappa \\
			\midrule
			Researcher v.s. Citizen & .80 \\
			Researcher v.s. Filtered MTurk workers & .81$\pm$.01 \\
			Researcher v.s. All MTurk workers & .75$\pm$.02 \\
			Citizen v.s. All MTurk workers & .72$\pm$.02 \\
			Citizen v.s. Filtered MTurk workers & .75$\pm$.01 \\
			\bottomrule
		\end{tabular}
		\caption{The inter-rater agreement (simulated 100 times) between pairs of MTurk workers, citizen, and researcher groups. The reported format is \quo{mean$\pm$standard deviation.}}
		\label{tab:kappa}
	\end{table}
	
	The data quality between citizens and filtered MTurk workers is similar.
	Filtered workers' page acceptance rates are better than 0.3 (random guessing is 0.07).
	To match the labeling tool's quality-control mechanism, we randomly selected three assignments for majority voting and simulated 100 times.
	Using researcher labels as the ground truth, Table~\ref{tab:mturk_citizen_compare} indicates similar strong performance of the positive labels (with smoke).
	The strong Cohen's kappa in Table~\ref{tab:kappa} shows high inter-rater agreement.
	
	\section{Experiments}
	
	\begin{table}[t]
		\centering
		\footnotesize
		\begin{tabular}{lcccccc}
			\toprule
			& $S_0$ & $S_1$ & $S_2$ & $S_3$ & $S_4$ & $S_5$ \\
			\midrule
			Training & .62 & .62 & .62 & .61 & .62 & .62 \\
			Validation & .13 & .12 & .12 & .14 & .11 & .13 \\
			Test & .25 & .26 & .26 & .25 & .26 & .25 \\
			\bottomrule
		\end{tabular}
		\caption{Ratio of the number of videos for each split (rounded to the nearest second digit). Split $S_0$,  $S_1$,  $S_2$,  $S_4$,  $S_5$ is based on views. Split $S_3$ is based on time sequence.}
		\label{tab:split}
	\end{table}
	
	We split the dataset into training, validation, and test sets in six ways ($S_0$ to $S_5$ in Table~\ref{tab:split}).
	Most of the splits (except $S_3$) are based on camera views, where different views are used for each split.
	Besides $S_3$, the splitting strategy is that each view is in the test set at least once, where 15 views from one site are distributed among the three sets, and four from two other sites are always in the test set.
	In this way, we can estimate whether the model is robust across views, as classification algorithms can suffer from overfitting in static cameras~\cite{beery2018recognition}.
	Split $S_3$ is based on time sequence, where the farthermost 18 days are used for training, the middle two days for validation, and the nearest 10 days for testing.
	Thus, we can evaluate whether models trained with data in the past can be used in the future.
	
	We treat our task as an action recognition problem and establish a baseline by using I3D ConvNet architecture with Inception-v1 layers~\cite{carreira-2017-quo}, a representative model for action recognition.
	The inputs of this baseline are RGB frames.
	The model is pretrained on ImageNet~\cite{russakovsky-2015-imagenet} and Kinetics~\cite{kay-2017-kinetics} datasets, and then fine-tuned on our training set.
	During training, we apply standard data augmentation, including horizontal flipping, random resizing and cropping, perspective transformation, area erasing, and color jittering.
	We refer to this baseline as RGB-I3D.
	The validation set is used for hyper-parameter tuning.
	Our baseline models are optimized using binary cross-entropy loss and Stochastic Gradient Descent with momentum 0.9 for 2,000 steps.
	Table~\ref{tab:experiment-parameters} shows the details of hyper-parameters. 
	All models and scripts are implemented in PyTorch~\cite{paszke-2017-automatic}.
	Model selection is based on F-score and validation error.
	
	\begin{table}[t]
		\centering
		\footnotesize
		\begin{tabu}{lccccccc}
			\toprule
			Model & $S_0$ & $S_1$ & $S_2$ & $S_3$ & $S_4$ & $S_5$ & Mean \\
			\midrule
			RGB-I3D & .80 & .84 & .82 & .87 & .82 & .75 & .817 \\
			\midrule
			RGB-I3D-ND & .76 & .79 & .81 & .86 & .76 & .68 & .777 \\
			RGB-SVM & .57 & .70 & .67 & .67 & .57 & .53 & .618 \\
			\midrule
			RGB-I3D-FP & .76 & .81 & .82 & .87 & .81 & .71 & .797 \\
			Flow-I3D & .55 & .58 & .51 & .68 & .65 & .50 & .578 \\
			Flow-SVM & .42 & .59 & .47 & .63 & .52 & .47 & .517 \\
			\midrule
			RGB-TSM & .81 & .84 & .82 & .87 & .80 & .74 & .813 \\
			RGB-LSTM & .80 & .84 & .82 & .85 & .83 & .74 & .813 \\
			\rowfont{\color{red}}
			\tikzmark{start}RGB-NL & .81 & .84 & .83 & .87 & .81 & .74 & .817\tikzmark{end} \\
			RGB-TC & .81 & .84 & .84 & .87 & .81 & .77 & .823 \\
			\bottomrule
		\end{tabu}
		\caption{F-scores for comparing the effect of data augmentation and temporal models on the test set for each split. Abbreviation ND and FP means no data augmentation and with frame perturbation, respectively. \textcolor{red}{There is an error (indicated in red), which is corrected in the Appendix.}}
		\label{tab:experiment}
		\tikz[remember picture] \draw[overlay] ([yshift=.35em]pic cs:start) -- ([yshift=.35em]pic cs:end);
	\end{table}
	
	In order to understand the effectiveness of our baseline, we also train five other models for comparison.
	RGB-I3D-ND is the same baseline model without data augmentation.
	RGB-SVM exploits Support Vector Machine (SVM) as the classifier, which takes the pretrained I3D features (without fine-tuning on our dataset) as input.
	RGB-I3D-FP is trained using video clips with frame-wise random permutation.
	Flow-I3D has the same network architecture as our baseline, but processes precomputed TVL1 optical flow frames.
	It also conducts the same data augmentation pipeline, except color jittering.
	Flow-SVM, similar to RGB-SVM, uses raw I3D features extracted with optical flow frames.
	
	Implementations of these five models are the same as mentioned.
	The ConvNets not noted in Table~\ref{tab:experiment-parameters} use the same hyper-parameters as RGB-I3D.
	Table~\ref{tab:experiment} shows that I3Ds outperform SVMs by a large margin, and data augmentation can improve the performance.
	Also, permuting frame ordering does not degrade the performance much, and the flow-based models perform worse than their RGB counterparts.
	To further understand the challenge of using temporal information, we train the other five variations based on RGB-I3D with different temporal processing techniques.
	RGB-NL wraps two Non-Local blocks~\cite{wang2018non} in the last Inception layer (closest to the output) around the 3D convolution blocks with a kernel size larger than one.
	RGB-LSTM attaches one Long Short-Term Memory layer~\cite{hochreiter1997long} with 128 hidden units after the last Inception layer.
	RGB-TSM wraps Temporal Shift modules~\cite{lin2019tsm} around each Inception layer.
	RGB-TC attaches one Timeception layer~\cite{hussein2019timeception} after the last Inception layer, using a 1.25 channel expansion factor.
	This variation is our best baseline model (Table~\ref{tab:best-result}).
	We fine-tune the LSTM and TC variation from the best RGB-I3D model with the I3D layers frozen.
	Table~\ref{tab:experiment} shows that these temporal processing techniques do not outperform the baseline model, which confirms the challenge of using the temporal information.
	
	\begin{table}[t]
		\centering
		\footnotesize
		\begin{tabular}{lccccc}
			\toprule
			Model
			& $\eta$
			& \begin{tabular}{@{}c@{}} Weight \\ decay \end{tabular}
			& $i$
			& \begin{tabular}{@{}c@{}} Batch \\ size \end{tabular}
			& Milestones \\
			\midrule
			RGB-I3D & $0.1$ & $10^{-6}$ & $2$ & $40$ & $(500, 1500)$ \\
			RGB-TSM & $0.1$ & $10^{-10}$ & $1$ & $40$ & $(1000, 2000)$ \\
			RGB-LSTM & $0.1$ & $10^{-4}$ & $1$ & $32$ & $(1000, 2000)$ \\
			RGB-TC & $0.1$ & $10^{-6}$ & $1$ & $32$ & $(1000, 2000)$ \\
			\bottomrule
		\end{tabular}
		\caption{Hyper-parameters. Symbol $\eta$ is the initial learning rate, and $i$ is the number of iterations to accumulate gradients before backward propagation. Milestones are the steps to decrease the learning rate by a factor of 0.1.}
		\label{tab:experiment-parameters}
	\end{table}
	
	\begin{table}[t]
		\centering
		\footnotesize
		\begin{tabular}{lccccccc}
			\toprule
			Metric & $S_0$ & $S_1$ & $S_2$ & $S_3$ & $S_4$ & $S_5$ & Average \\
			\midrule
			Precision & .87 & .84 & .92 & .88 & .88 & .78 & .862 \\
			Recall & .76 & .83 & .77 & .87 & .76 & .76 & .792 \\
			F-score & .81 & .84 & .84 & .87 & .81 & .77 & .823 \\
			ROC/AUC & .90 & .94 & .94 & .95 & .92 & .91 & .927 \\
			\bottomrule
		\end{tabular}
		\caption{Evaluation of the best baseline model (RGB-TC) on the test set for each split. ROC/AUC means area under the receiver operating characteristic curve.}
		\label{tab:best-result}
	\end{table}
	
\begin{figure}[t]
		\centering
		\begin{tabular}{c}
			\includegraphics[width=0.9\columnwidth]{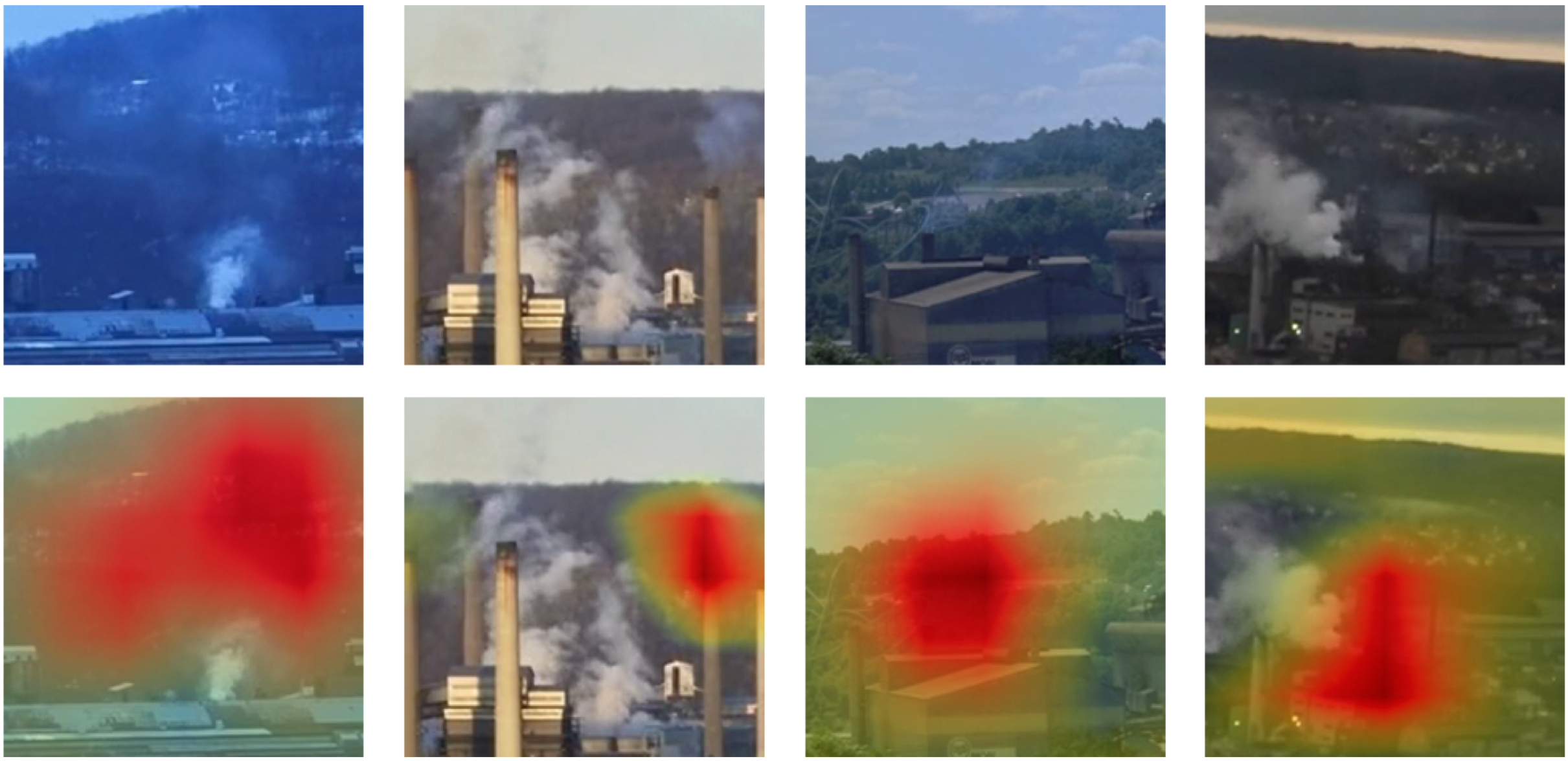}
		\end{tabular}
		\caption{True positives in the test set from split $S_0$. The top and bottom rows show the original video frame and the overlaying heatmap of Class Activation Mapping, respectively.}
		\label{fig:grad-cam-good}
	\end{figure}

	To check how our baseline model makes decisions, we visualize the semantic concept by applying Gradient-weighted Class Activation Mapping~\cite{selvaraju-2017-grad}.
	It uses the gradients that flow into the last convolutional layer to generate a heatmap, highlighting areas that affect the prediction.
	Ideally, our model should focus on smoke emissions instead of other objects in the background ({\em e.g.}, stacks or facilities).
	Figure~\ref{fig:grad-cam-good} shows true positives (sampled from 36 frames) for smoke and co-existence of both smoke and steam.
	
	\section{Survey Study}
	
	\begin{table}[t]
		\centering
		\footnotesize
		\begin{tabular}{lccccccc}
			\toprule
			Age & 25-34$\!\!$ & 35-44$\!\!$ & 45-54$\!\!$ & 55-64$\!\!$ & 65+$\!\!$ & Unknown$\!\!$  \\
			\midrule
			& 1 & 2 & 2 & 1 & 4 & 1 \\
			\bottomrule
		\end{tabular}
		\caption{Demographics of 11 survey responses. Four are female. Eight have college degrees (or above).}
		\label{tab:demographics}
	\end{table}
	
	We conducted a survey to learn how and why volunteers participate.
	The survey was online (Google Forms), anonymous, voluntary, and had no compensation.
	We distributed the survey through community leaders and via email lists that we curated during the design workshops.
	We collected 11 responses.
	Table~\ref{tab:demographics} shows the demographics of the survey participants.
	Most of them labeled smoke emissions (82\%) and discussed the project in person with others (82\%).
	
	We asked an open-ended question to identify motivations.
	Selected quotes are in parentheses.
	Three participants noted the importance in advocacy (\textit{\quo{It is an essential tool for improving regional air quality}}), and two mentioned the desire to help research (\textit{\quo{To help provide more data for researchers to work with [...]}}).
	Others referred to their background (\textit{\quo{I have been interested in air pollution from local steel mills since childhood and have wanted to do something about it}}), a desire to push regulators (\textit{\quo{[...] it is my hope that we can use that information to force the regulatory agencies to do a better job of enforcing clean air regulations}}), and a wish to adopt our system (\textit{\quo{[...] hopes that this technology can help in our monitoring of other industries [...]}}).
	
	Although participants believed our system can raise public awareness of pollution (\textit{\quo{[...] wider marketing to community members not already involved/concerned with AQ initiatives}}), there was a social-technical gap~\cite{ackerman2000the} between community expectations and the system's capability.
	For example, in another open-ended question about community feedback, participants express curiosity about the correlation between smoke emissions and air quality complaints (\textit{\quo{[...] when there are more visible emissions, are there also more reports of smells?}}).
	Also, participants expect the information to be useful to policy-makers (\textit{\quo{The videos may even provide motivation for local politicians to revise emissions regulations in a way that makes them more strict}}).
	Based on the survey study, more co-design workshops are needed to create data-driven evidence for making social impact, including policy changes.
	
	\section{Discussion}
	
	\textbf{Community Engagement Challenges.}
	Citizen science can empower residents to address air pollution~\cite{hsu-2019-smell,hsu-2017}, but engaging communities is challenging.
	For example, setting up cameras and recruiting volunteers requires substantial community outreach efforts.
	Communities who suffer from air pollution are often financially impoverished, whose top priority might not be improving air quality.
	Moreover, air pollution issues are not as enjoyable as the topics of other citizen science projects in astronomy~\cite{lintott-2008-galaxy} or bird watching~\cite{sullivan-2009-ebird}.
	Also, labeling industrial smoke emissions requires volunteers to understand smoke behavior and is thus harder than recognizing generic objects.
	We tackled the challenges by adopting community co-design, where the people who are impacted the most participate in the design process. Understanding community expectations and concerns through conversations ({\em e.g.}, local meetings or workshops) is the key to building mutual trust and sustaining community-powered projects.
	
	\textbf{Wicked Problems in Data Collection.}
	Community-powered projects like \dataset suffer from the dilemma of ``Wicked Problems''~\cite{rittel1973dilemmas}: \textit{they have no precise definition, cannot be fully observed initially, depend on context, have no opportunities for trial and error, and have no optimal or provably correct solutions.}
	In this case, multiple uncontrollable factors can affect the data and label distribution.
	For instance, the availability of human labeling power within citizen groups varies over time, as community agenda changes often.
	Moreover, our system co-design and data labeling processes happened simultaneously and iteratively, which meant the user interface changed at each design stage, and the amount of available data was incremental.
	Also, we have no prior knowledge about the distribution of smoke emission across time since obtaining such information requires having profound knowledge about the pollution sources' working schedule, which is not available.
	The community dynamics and underlying uncertainty can lead to imbalanced datasets.
	One could pursue crowdsourcing for better data quality, but crowdsourcing relinquishes the opportunity to build informed communities that can take action for social change.
	Making AI systems work under such real-world constraints remains an open research question.
	
	\textbf{Integration of Citizen Science and Crowdsourcing.}
	Our analysis finds performance between volunteers and MTurk workers to be similar, suggesting opportunities for combining citizen science and crowdsourcing due to their comparable reliability.
	For example, at the early stage of development, one could recruit MTurk workers to label an initial small-scale dataset.
	Once the model is trained with the initial dataset, one can use it to produce visual evidence that can attract active citizen scientists.
	At this stage, the model accuracy may not reach the desired level, and human intervention may be required.
	However, as the community's attention starts to increase, it would be possible to collect more labels and improve model performance over time.
	Also, as the uncertainty in community dynamics can lead to imbalanced datasets, crowdsourcing can be a way to improve the label distribution.
	We leave this to future work.
	
	\textbf{Limitations.}
	Project \dataset tackles Wicked Problems, which means we can only make practical design decisions based on available information at each design stage.
	These practical but not optimal decisions lead to several limitations.
	Comparing the reliability between citizens and MTurk workers may be unfair due to the difference in their label-aggregation logic.
	Also, models trained on our 19 views from three facilities might not generalize well to other industrial facilities.
	Furthermore, \dataset does not include nighttime videos, which are difficult to label due to insufficient light.
	Moreover, our dataset does not offer bounding box labels.
	We applied domain knowledge to define the locations that smoke emissions were likely to occur, making smoke recognition a classification rather than a detection problem.
	Finally, there are other ways for aggregating labels provided by researchers and citizens, such as EM-based methods~\cite{raykar2010learning}.
	In our case, researchers always overrode the decisions made by citizens.
	We leave the expansion of different label types and the methodology for aggregation decisions from various user groups to future work.
	
	\section{Conclusion}
	
	Project \dataset shows that, besides model performance, AI systems' social impact and ethics are also critical.
	We hope to reinforce citizens' voices and rebalance power relationships among stakeholders through system design and deployment.
	We have deployed the AI model to recognize smoke.
	Community activists and health department officers are working with our system to curate a list of severe pollution events as evidence to conduct air pollution studies.
	We envision that our work can encourage others to keep communities in the center of every AI system design stage.
	Communities affected most by social or environmental problems know their needs best and have active roles in our project.
	By adopting community co-design, our work demonstrates a way to forge sustainable alliances and shared prosperity between academic institutions and local citizen groups.
	
	\section{Acknowledgments}
	
	We thank GASP (Group Against Smog and Pollution), Clean Air Council, ACCAN (Allegheny County Clean Air Now), Breathe Project, NVIDIA, and the Heinz Endowments for the support of this research. We also greatly appreciate the help of our volunteers, which includes labeling videos and providing feedback in system development.
	
	\bibliography{reference}
	
	\section{Appendix}
	
	There was an error in the code when we implemented the RGB-NL model with Non-Local blocks, which means that the RGB-NL model performance in Table~\ref{tab:experiment} is wrong. We fixed the code and ran the experiment for the RGB-NL model again. The corrected result is in Table~\ref{tab:experiment-correct}.
	
	\begin{table}[h]
		\centering
		\footnotesize
		\begin{tabu}{lccccccc}
			\toprule
			Model & $S_0$ & $S_1$ & $S_2$ & $S_3$ & $S_4$ & $S_5$ & Mean \\
			\midrule
			RGB-NL & .80 & .84 & .83 & .86 & .82 & .75 & .817 \\
			\bottomrule
		\end{tabu}
		\caption{Correction of the RGB-NL model performance.}
		\label{tab:experiment-correct}
	\end{table}
	
\end{document}